\documentclass{article}
\usepackage{iclr2027_conference,times}

\usepackage{amsmath,amsfonts,bm}

\def\eqref#1{equation~\ref{#1}}

\def\1{\bm{1}}

\DeclareMathAlphabet{\mathsfit}{\encodingdefault}{\sfdefault}{m}{sl}
\SetMathAlphabet{\mathsfit}{bold}{\encodingdefault}{\sfdefault}{bx}{n}

\usepackage{amsmath}
\usepackage{amssymb}
\usepackage{amsthm}
\usepackage{xcolor}
\usepackage{booktabs}
\usepackage{graphicx}
\usepackage{multirow}
\usepackage{microtype}
\usepackage{hyperref}
\usepackage{url}
\usepackage{enumitem}
\usepackage{algorithm}
\usepackage{algpseudocode}
\usepackage{flafter}
\usepackage{xspace}

\title{SchurQuant: Groupwise Discrete Optimization for Layer-Wise LLM Quantization}

\author{
Gunjun Lee
\And
Sehwan Son
\And
Younjoo Lee
\And
Byungjun Kim
\And
Jung Ho Ahn
\AND
Seoul National University \\
Seoul, South Korea \\
\texttt{\{kevin970401,sonsh011106,younjoo0614,bjkim1023,gajh\}@snu.ac.kr}
}
\iclrfinalcopy

\newcommand{\method}{\textsc{SchurQuant}\xspace}
\newcommand{\optimizer}{\textsc{SchurOpt}\xspace}
\newcommand{\Wref}{W^{\mathrm{ref}}}
\newcommand{\Xref}{X^{\mathrm{ref}}}
\newcommand{\Qset}{\mathcal{Q}}
\newcommand{\tr}{\operatorname{tr}}
\newcommand{\clip}{\operatorname{clip}}
\newcommand{\Wfixed}{\ensuremath{W_F}}
\newcommand{\Wcurrent}{\ensuremath{W_c}}
\newcommand{\Wremaining}{\ensuremath{W_r}}
\algrenewcommand\algorithmicrequire{\textbf{Input:}}
\algrenewcommand\algorithmicensure{\textbf{Output:}}

\newtheorem{proposition}{Proposition}

\begin{document}
\maketitle \footnotetext{Correspondence to: Gunjun Lee \textless kevin970401@snu.ac.kr\textgreater, Jung Ho Ahn \textless gajh@snu.ac.kr\textgreater}
\begin{abstract}
Weight-only post-training quantization (PTQ) enables the deployment of large language models under tight memory budgets, but accuracy often collapses at 2--3 bits. 
Existing backpropagation-free PTQ optimizers have two limitations: group decisions ignore the correction that the remaining continuous suffix can absorb, and discrete refinements typically keep the affine quantization grid fixed. 
We introduce \optimizer, which analytically eliminates the suffix's optimal continuous response, yielding an exact groupwise quadratic with Schur-complement curvature.
It then alternates closed-form row-wise scale/zero-point refitting with coordinate descent over integer codes. With the GPTQ objective fixed, \optimizer improves mean zero-shot accuracy on 2-bit Qwen3-4B by 11.88 percentage points (pp). 
At higher precision, however, tighter reconstruction does not consistently improve end-model metrics.
\method therefore combines \optimizer with quantized-prefix teacher reconstruction, reference-weight regularization, residual-add targets, and teacher-decision token weighting. 
Across eight Llama and Qwen models, \method achieves the highest mean zero-shot accuracy among the evaluated backpropagation-free PTQ baselines, outperforming the strongest baseline by 9.65 pp at 2 bits.
\end{abstract}

\section{Introduction}
\label{sec:introduction}

Deploying large language models (LLMs) is increasingly constrained by memory capacity and bandwidth rather than arithmetic throughput~\citep{arxiv-2026-kimik3}, and these constraints are most severe on edge devices. 
Weight-only post-training quantization (PTQ) addresses this by converting pretrained weights to low precision using only a small calibration set without retraining.
Yet conventional weight-only PTQ collapses sharply in perplexity and downstream accuracy in 2--3-bit quantization.

Weight-only PTQ can be decomposed into two largely independent components. 
The \emph{objective} determines which error is minimized, classically the output-reconstruction loss $\lVert W X - \Wref X\rVert_F^2$ of a linear layer, where $\Wref$ is the teacher weight, $X$ contains calibration activations, and $W$ is the quantized weight.
These activations determine the curvature $XX^\top$ and thus which weight-error directions matter (Appendix~\ref{sec:calibration}).
Recent methods enrich this objective with activation errors accumulated in preceding layers, asymmetric calibration, joint correction of past error and future propagation, residual-path reconstruction, or combinations of relaxed block objectives~\citep{neurips-2025-qep,icml-2025-gptaq,iclr-2026-qronos,arxiv-2026-loaq,arxiv-2025-lpcd}.

The \emph{optimizer} determines how that objective is solved on an integer grid. 
GPTQ visits weights along the input-column axis, rounds each to a nearby level, and propagates the induced error to the remaining full-precision weights through an inverse Hessian~\citep{iclr-2023-gptq}.
This update is equivalent to Babai's nearest-plane algorithm on the lattice defined by $XX^\top$~\citep{iclr-2026-gptqbabai}.
More recent methods use cyclic or greedy coordinate descent to refine discrete weights or groups~\mbox{\citep{arxiv-2023-quantease,arxiv-2024-cdquant,arxiv-2025-lpcd}}.
GuidedQuant additionally injects end-loss gradients into the layer-wise objective and derives a monotone scalar quantizer~\citep{icml-2025-guidedquant}. 

The missing step is to condition a group decision on variables that remain adjustable. 
Prior coordinate descent evaluates the group under the raw block $G_{cc}$ of a symmetric curvature matrix $G$, effectively holding the continuous suffix fixed, although that suffix can absorb part of the error.
The exact curvature remaining after its best continuous response is the Schur complement $S=G_{cc}-G_{cr}G_{rr}^{-1}G_{rc}$, which can differ substantially from $G_{cc}$ under strong coupling.

Based on this observation, we introduce \optimizer, the Schur-conditioned discrete optimizer, and refer to its pairing with our final augmented surrogate as \method.
We partition the input dimension into chunks of one quantization group.
While quantizing a current chunk \Wcurrent{}, the already quantized prefix \Wfixed{} is held fixed and the suffix \Wremaining{} remains continuous.
Analytically minimizing over \Wremaining{} produces $S$ and a corresponding conditional linear term. 
\optimizer then optimizes the group against this reduced quadratic: it jointly refits each row's scale and zero-point for fixed codes and updates codes by coordinate descent for fixed grid parameters. 

Holding the objective exactly equal to GPTQ's isolates the optimizer's contribution.
On Qwen3-4B with 2-bit codes, \optimizer reduces WikiText-2 perplexity from 258.96 to 81.95 and improves mean zero-shot accuracy by 11.88 pp.
At 3 and 4 bits, however, the same optimizer worsens perplexity while changing mean zero-shot accuracy only marginally.
This reveals a second bottleneck.
Layer-wise reconstruction is a loose surrogate for the final language-model loss.
Motivated by this mismatch, our final configuration combines quantized-prefix teacher reconstruction with a reference-weight anchor, a residual-add target, and teacher-decision token weighting. The last term upweights the calibration positions whose clean top-1 prediction has already been changed by the quantized prefix.

Our contributions are as follows:
\begin{itemize}[leftmargin=*]
    \item We derive the exact discrete group problem conditioned on a fixed prefix and the optimal continuous suffix response, then optimize its $(S,T)$ with a joint closed-form scale/zero-point refit and row-parallel coordinate descent. This conditioning, rather than coordinate descent itself, is the optimizer contribution.
    \item We integrate prior reference-weight regularization and residual-add reconstruction into the sufficient statistics and introduce teacher-decision token weighting, changing the surrogate without changing \optimizer.
    \item We separate optimizer and objective effects experimentally. Across eight Llama and Qwen models under 2--4-bit quantization, \method achieves the highest mean zero-shot accuracy among the evaluated backpropagation-free PTQ baselines, with its largest advantage at 2 bits.
\end{itemize}

\section{Background and Motivation}
\label{sec:background}

\subsection{Uniform affine quantization}
\label{sec:affine-quantization}

A uniform affine quantizer maps a real weight $w$ to an integer code $z$ as
\begin{equation}
    z = \clip\!\left(\left\lfloor \frac{w}{a}\right\rceil + o,
    q_{\min},q_{\max}\right),
    \qquad \widehat w = a(z-o),
    \label{eq:affine-quantize}
\end{equation}
where $a>0$ is the spacing between adjacent levels, $o$ is the code corresponding to zero, $\lfloor\cdot\rceil$ denotes nearest-integer rounding, and clipping enforces the representable range. 
With unsigned $b$-bit codes, $q_{\min}=0$ and $q_{\max}=2^b-1$. 
The symmetric case fixes $o=0$ and uses signed codes $q_{\min}=-2^{b-1}$ and $q_{\max}=2^{b-1}-1$.
The representable set is $\Qset(a,o)=\{a\left(\ell-o\right):\ell\in\{q_{\min},\ldots,q_{\max}\}\}$.

For a weight matrix in $\mathbb{R}^{d_{\mathrm{out}}\times d_{\mathrm{in}}}$, per-row group quantization divides each row along the input dimension into groups of size $g$ and assigns each row--group its own quantizer.
We call the $d_{\mathrm{out}}\times g$ submatrix processed at one step a \emph{chunk}. 
An asymmetric quantizer stores an FP16 scale and a $b$-bit zero-point for each group.
With FP16 metadata, this gives an effective weight precision of approximately $b+16/g$ bits. 
All experiments use $g=128$, the standard weight-only PTQ setting introduced with GPTQ, which limits metadata overhead to approximately $0.125$ bits per weight. 

\subsection{GPTQ, propagated-error objectives, and discrete refinement}
\label{sec:gptq-qep}

GPTQ minimizes the output-reconstruction objective $\min_{W\in\Qset^{d_{\mathrm{out}}\times d_{\mathrm{in}}}}\lVert \Wref X-WX\rVert_F^2$ of each linear layer~\citep{iclr-2023-gptq}. It initializes a uniform grid for each row--group, visits columns sequentially, and rounds the current weights. To reduce the output effect of the rounding error, it updates the remaining full-precision weights using the inverse input Hessian $H=XX^\top$
\begin{equation}
    \delta_{>q}
    =-\frac{w_q-\operatorname{quant}(w_q)}{[H^{-1}]_{qq}}
      [H^{-1}]_{q,>q}.
    \label{eq:gptq-update}
\end{equation}
All output rows share the column update and can be processed in parallel. Crucially, the discrete decision itself remains nearest-level rounding; curvature enters only through the correction applied afterward. The equivalence to Babai's nearest-plane algorithm makes this limitation explicit~\citep{iclr-2026-gptqbabai}.
Here the subscript $>q$ denotes input columns scheduled after column $q$.

GPTQ is operation-local, not accounting for activation error produced by already quantized operations. 
Let $\Xref_\ell$ and $X_\ell$ denote the inputs to operation $\ell$ produced by the clean and already quantized prefixes, respectively. QEP~\citep{neurips-2025-qep} optimizes the current quantized weight on $X_\ell$ to reproduce the clean teacher output $\Wref_\ell\Xref_\ell$:
\begin{equation}
    \min_{W_\ell\in\Qset^{d_{\mathrm{out}}\times d_{\mathrm{in}}}}
    \left\lVert \Wref_\ell \Xref_\ell-W_\ell X_\ell\right\rVert_F^2,
    \qquad \delta_\ell:=\Xref_\ell-X_\ell.
    \label{eq:qep-objective}
\end{equation}

The optimum of its continuous relaxation is a corrected teacher weight, after which QEP uses the same GPTQ-style discrete optimizer.
FOEM retains the first-order term arising away from a stationary pretrained weight~\citep{aaai-2026-foem}.
LoaQ further extends the reconstruction scope from linear-layer outputs to sub-layer outputs by explicitly matching the residual-added hidden state, thereby accounting for error accumulated along the residual stream while retaining the layer-wise PTQ formulation~\citep{arxiv-2026-loaq}. 
These methods, and work on asymmetric calibration or error propagation~\citep{icml-2025-gptaq,iclr-2026-qronos}, primarily change the objective or reconstruction scope rather than how a group is conditioned on its continuous suffix.

Coordinate optimization of quantized parameters is already used. 
QuantEase uses cyclic coordinate descent for the discrete layer-reconstruction problem and proves non-increase of its objective~\citep{arxiv-2023-quantease}.
CDQuant directly contrasts greedy coordinate descent with GPTQ's predetermined pass and extends it to group quantization~\citep{arxiv-2024-cdquant}.
To our knowledge, their discrete updates do not first analytically eliminate the optimal response of the continuous suffix.
\optimizer differs at this conditioning step. 
It applies both parameter refitting and code updates to $S=G_{cc}-G_{cr}G_{rr}^{-1}G_{rc}$ and the corresponding reduced linear term.

\subsection{Schur complement}
\label{sec:schur-background}

The Schur complement~\citep{doi-1917-schur} measures the curvature that remains after adjustable variables make their optimal response. 
Let
\begin{equation}
x=
\begin{bmatrix}
x_c\
x_r
\end{bmatrix}
\end{equation}
denote a generic optimization variable partitioned into a current block $x_c$ and a remaining adjustable block $x_r$.
Partition a symmetric curvature matrix as
\begin{equation}
    G=\begin{bmatrix}G_{cc}&G_{cr}\\G_{rc}&G_{rr}\end{bmatrix},
    \qquad G_{rc}=G_{cr}^\top.
\end{equation}
For the quadratic form induced by $G$, if $G_{rr}$ is invertible then minimizing over $x_r$ gives $x_r^\star=-G_{rr}^{-1}G_{rc}x_c$ and
\begin{equation}
    \min_{x_r}f(x_c,x_r)
    =\tfrac12x_c^\top
      \underbrace{\left(G_{cc}-G_{cr}G_{rr}^{-1}G_{rc}\right)}_{G/G_{rr}}
      x_c.
    \label{eq:schur-definition}
\end{equation}
The subtracted term is the part that $x_r$ can absorb, so $G/G_{rr}\preceq G_{cc}$ whenever $G\succeq0$. In quantization, the current group corresponds to $x_c$ and the continuous suffix to $x_r$; the Schur complement therefore gives the exact local curvature after the best possible suffix correction.

\subsection{Related quantization approaches}
\label{sec:related-work}

\paragraph{Quantization-aware optimization.}
Quantization-aware training (QAT) uses fake-quantized weights and straight-through gradients to update weights or quantizer parameters.
EfficientQAT trains weights within each Transformer block and then optimizes quantization parameters end to end~\citep{acl-2025-efficientqat}. 
Although accurate, such training requires repeated forward and backward passes and is substantially more expensive than gradient-free layer-wise PTQ.

\paragraph{Calibration-optimized PTQ.}
OmniQuant learns clipping parameters and equivalent transformations through blockwise reconstruction on calibration data~\citep{iclr-2024-omniquant}, whereas AWQ identifies salient weight channels from activation statistics and protects them through per-channel scaling~\citep{mlsys-2024-awq}. These are more naturally classified as PTQ: they adapt the objective or representation from calibration data rather than retraining the model end to end. \method{} is complementary, as it changes the discrete optimizer applied after the layer-wise objective has been specified.

A second line changes the representation rather than the search. Rotation-based methods apply function-preserving basis changes to spread outliers across channels, using randomized Hadamard rotations~\citep{neurips-2024-quarot} or the rotations learned against a calibration loss~\citep{iclr-2025-spinquant}. 
Vector-quantized methods map groups of weights to codewords, combining incoherence processing with lattice codebooks~\citep{neurips-2023-quip,icml-2024-quipsharp} or learning additive codebooks~\citep{icml-2024-aqlm}. 
Both families change the inference path, requiring online rotations or packed-index decoding and lookup. In contrast, \method{} retains the same scalar per-row group format as GPTQ; only the offline optimizer changes.

\section{\optimizer{} and \method{}}
\label{sec:method}

We first define the quadratic sufficient-statistics interface shared by both methods: \optimizer and \method. Sections~\ref{sec:schur-objective}--\ref{sec:code-sweep} derive \optimizer, and Section~\ref{sec:align-objective} augments the objective to obtain the complete \method configuration.

\subsection{Problem formulation}
\label{sec:problem-formulation}

Consider one linear projection with weights $W\in\mathbb{R}^{d_{\mathrm{out}}\times d_{\mathrm{in}}}$, calibration inputs $X=[x_1,\ldots,x_N]\in\mathbb{R}^{d_{\mathrm{in}}\times N}$, and teacher targets $Y=[y_1,\ldots,y_N]\in\mathbb{R}^{d_{\mathrm{out}}\times N}$.
Let $\Omega=\operatorname{diag}(\omega_1,\ldots,\omega_N)$ contain nonnegative token weights.
Throughout this paper, $\Wref$ denotes the clean pretrained weight and $W$ the student working weight being discretized; after quantization, $W\in\Qset^{d_{\mathrm{out}}\times d_{\mathrm{in}}}$. During a group update, the fixed prefix $W_F$ and candidate current group $W_c$ are discrete, whereas the remaining suffix $W_r$ is optimized as a continuous variable. We refer to $W_r$ as the \emph{continuous suffix}. Unless a controlled GPTQ-objective experiment states otherwise, $X$ is collected from the current quantized-prefix stream and $Y$ from the clean teacher stream.
We optimize
\begin{equation}
    \mathcal{L}(W)
    =\frac12\lVert (WX-Y)\Omega^{1/2}\rVert_F^2
     +\frac{\lambda}{2}\lVert W-\Wref\rVert_F^2,
    \label{eq:general-objective}
\end{equation}
where the dimensionless $\lambda$ anchors the quantized solution to the pretrained weights.
Expanding Eq.~\ref{eq:general-objective} gives
\begin{align}
    G&=X\Omega X^\top+\lambda I,
    &C&=Y\Omega X^\top+\lambda\Wref, \label{eq:sufficient-statistics}
\end{align}
\begin{align}
    \mathcal{L}(W)
    &=\tfrac12\tr(WGW^\top)-\tr(CW^\top)+\mathrm{const}.
    \label{eq:quadratic-objective}
\end{align}
Thus $(G,C)$ are sufficient statistics: the optimizer need not retain $X$, $Y$, or $\Omega$. 
If $\lambda>0$, $G$ is positive definite even when $X\Omega X^\top$ is rank deficient.

Eq.~\ref{eq:quadratic-objective} is also the interface between \optimizer{} and a layer-wise objective.
Setting $\Omega=I$, $\lambda=0$, and $Y=\Wref X$ recovers GPTQ.
Using quantized-prefix inputs with clean teacher outputs recovers the QEP-style asymmetric teacher-reconstruction objective.
Residual-add targets and token weights only change how $(G,C)$ are collected; the optimizer below is unchanged.

\subsection{The Schur-reduced group objective}
\label{sec:schur-objective}

For any candidate $W_c$, the suffix can still react before it is quantized. 
We therefore eliminate its optimal continuous response,
\begin{equation}
    W_c^{\star,\mathrm{Schur}}
    =\arg\min_{W_c\in\Qset}\ \mathcal{L}\!\left(W_c,W_r^\star(W_c)\right),
    \qquad
    W_r^\star(W_c)=\arg\min_{W_r}\mathcal{L}(W_c,W_r),
    \label{eq:schur-discrete-problem}
\end{equation}
so that the selected group minimizes the loss remaining after the best possible continuous suffix correction, not merely the loss of the group in isolation.

\begin{proposition}[Schur reduction]
\label{prop:schur}
Fix $W_F$ and assume that $G_{rr}$ is invertible.
Eliminating $W_r$ yields the equivalent local objective
\begin{align}
    \mathcal{L}_{\mathrm{local}}(W_c)
    &=\frac12\tr(W_cSW_c^\top)-\tr(TW_c^\top)+\mathrm{const},
    \label{eq:local-objective}\\
    S&=G_{cc}-G_{cr}G_{rr}^{-1}G_{rc},
    \qquad
    T=C_c^{\mathrm{eff}}-C_r^{\mathrm{eff}}G_{rr}^{-1}G_{rc},
    \label{eq:local-curvature}\\
    C_c^{\mathrm{eff}}&=C_c-W_FG_{Fc},
    \qquad\quad\ \,
    C_r^{\mathrm{eff}}=C_r-W_FG_{Fr}.
    \label{eq:effective-linear}
\end{align}
\end{proposition}

The proof is in Appendix~\ref{app:proof-schur}. 
Intuitively, $G_{cc}$ is the direct curvature of perturbing the current chunk, while $G_{cr}G_{rr}^{-1}G_{rc}$ is the part absorbable by the optimal continuous suffix, so $S\preceq G_{cc}$.
The linear term $T$ likewise accounts for both the error already introduced by the prefix and the component that the suffix can absorb.

Computing $S$ by factorizing every suffix $G_{rr}$ would be prohibitively expensive. 
Let t denote the starting input-column index of the current chunk. We maintain $P=(G_{[t:,t:]})^{-1}$ partitioned as $\left[\begin{smallmatrix}P_{cc}&P_{cr}\\P_{cr}^\top&P_{rr}\end{smallmatrix}\right]$ with $P_{cc}\in\mathbb{R}^{g\times g}$.
The block-inverse identity gives
\begin{equation}
    S=P_{cc}^{-1},
    \qquad G_{rr}^{-1}G_{rc}=-P_{cr}^\top S,
    \qquad P_{\mathrm{next}}=P_{rr}-P_{cr}^\top SP_{cr}.
    \label{eq:inverse-recurrence}
\end{equation}
Each chunk therefore requires one $g\times g$ inverse and an $O(m^2g)$ update for a suffix of length $m$, rather than a new $O(m^3)$ factorization.

\subsection{Exact row-wise quantizer-parameter update}
\label{sec:qparam-fit}

For output row $i$, write the quantized current group as $q_i = a_i(z_i - o_i\mathbf{1}^\top)$ with integer codes $z_i\in\{q_{\min},\ldots,q_{\max}\}^g$, scale $a_i$, and zero-bias $o_i$. A symmetric quantizer is the special case $o_i=0$. Because Eq.~\ref{eq:local-objective} separates across output rows,
\begin{equation}
    \mathcal{L}_{\mathrm{local}}(Q)
    =\sum_{i=1}^{d_{\mathrm{out}}}
      \left(\tfrac12q_iSq_i^\top-t_iq_i^\top\right),
    \label{eq:rowwise-objective}
\end{equation}
all rows can be updated in parallel.

\begin{proposition}[Row-wise scale and zero-point update]
\label{prop:qparam}
For fixed $z_i$ and a candidate $o$, define $u_i(o)=z_i-o\mathbf{1}^\top$,
$A_i(o)=u_i(o)Su_i(o)^\top$, and $d_i(o)=t_i u_i(o)^\top$. The optimal
positive scale and zero-point are
\begin{equation}
    a_i^\star(o)=\max\!\left(\frac{d_i(o)}{A_i(o)},\epsilon_a\right),
    \qquad
    o_i^\star=\arg\min_{o\in\{0,\ldots,2^b-1\}}
    \left[\frac12a_i^\star(o)^2A_i(o)-a_i^\star(o)d_i(o)\right].
    \label{eq:qparam-update}
\end{equation}
\end{proposition}
Only $2^b$ zero-points are considered, so the enumeration is inexpensive at the
target precisions. The required coefficients are obtained from one product
$ZS$, followed by row-wise reductions and evaluation of the $2^b$ candidates.
Appendix~\ref{app:proof-qparam} gives the proof.

\subsection{Coordinate descent over integer codes}
\label{sec:code-sweep}

Stack the row code vectors into $Z$, with scales $a\in\mathbb{R}_+^{d_{\mathrm{out}}}$ and integer zero-points $o\in\{0,\ldots,2^b-1\}^{d_{\mathrm{out}}}$. For fixed $(a,o)$, the current chunk is $Q=\operatorname{diag}(a)(Z-o\mathbf{1}^\top)$. At column $j$, row $i$ can choose among the $2^b$ candidates $v_{i,k}=a_i(k-o_i)$.
Retaining only terms that depend on $v$ yields the coordinatewise objective, up to an additive constant,
\begin{equation}
    \ell_{i,j}(v)
    =\frac12S_{jj}v^2
      +\left(
      \underbrace{(q_iS)_j-q_{i,j}S_{jj}}_{\operatorname{cross}_{i,j}}
      -t_{i,j}\right)v,
    \qquad
    z_{i,j}\leftarrow
    \arg\min_k \ell_{i,j}\!\left(a_i(k-o_i)\right).
    \label{eq:coordinate-objective}
\end{equation}
At the start of a sweep, we initialize $\Phi\leftarrow QS$. Maintaining this state makes $\operatorname{cross}_{i,j}=\Phi_{i,j}-q_{i,j}S_{jj}$ immediately available.
When the coordinate changes, define $\Delta_i=q_{i,j}^{\mathrm{new}}-q_{i,j}^{\mathrm{old}}$ and update the state by the rank-one operation $\Phi_{i,:}\leftarrow\Phi_{i,:}+\Delta_iS_{j,:}$.
Evaluating all $2^b$ candidates requires $\mathcal{O}(2^b)$ time and the state update $\mathcal{O}(g)$ per coordinate, for a total sweep complexity of $\mathcal{O}(d_{\mathrm{out}}g(2^b+g))$.

The cross term is the key difference from independent rounding.
It captures coupling with other columns in the group, while $S$ already accounts for correction available outside the group.
GPTQ reduces the curvature to a scalar but still makes a nearest-level decision.
\optimizer uses the reduced curvature both to fit the grid and to choose the integer codes themselves.

\subsection{Augmenting the local objective}
\label{sec:align-objective}

Even an exact optimizer can overfit a layer-wise surrogate. We therefore augment its sufficient statistics with a reference-weight anchor, a residual-add target, and teacher-decision token weighting while leaving \optimizer{} unchanged.

\paragraph{Reference-weight anchor.}
Following the regularized formulation of QEP~\citep{neurips-2025-qep}, we add
\[
    \frac{\lambda}{2}\lVert W-\Wref\rVert_F^2,
\]
which contributes $\lambda I$ to $G$ and $\lambda\Wref$ to $C$ in Eq.~\ref{eq:sufficient-statistics} without changing the optimizer.

\paragraph{Residual-add target.}
Following the sub-layer output approximation of LoaQ~\citep{arxiv-2026-loaq}, we use the clean residual output $Y=h+f(h)$ rather than only the projection-local output. Student-side statistics are collected from activations propagated through the quantized prefix. This changes the target-dependent sufficient statistics while leaving \optimizer{} unchanged.

\paragraph{Teacher-decision token weighting.}
Before quantizing Transformer layer $\ell$, we compare the logits $s_i^{Q,\ell}$ of the current quantized-prefix model with cached clean-teacher logits $s_i^T$. We define
\begin{equation}
    \tau_i=\arg\max_v s_{i,v}^T,
    \qquad
    m_i^{(\ell)}=\mathbf{1}\!\left[
       \arg\max_v s_{i,v}^{Q,\ell}\ne\tau_i
    \right],
    \qquad
    \alpha_i^{(\ell)}=1+(\rho-1)m_i^{(\ell)}.
    \label{eq:top-mismatch-weight}
\end{equation}
Thus a flipped top-1 prediction receives $\rho$ times the mass of an unchanged prediction. We use $\rho=8$ and normalize $\alpha_i^{(\ell)}$ to unit mean.

The mask is shared across projections in a Transformer layer.
For each projection, we compute the detached residual $r_i=\lVert \Wref x_i-y_i\rVert_2$ and use
\begin{equation}
    \omega_i=
    \frac{\bar\alpha_i^{(\ell)}}{\max(r_i,\epsilon)}
    \label{eq:effective-token-weight}
\end{equation}
in $\Omega$. The resulting statistics remain $X\Omega X^\top$ and $Y\Omega X^\top$ in Eq.~\ref{eq:sufficient-statistics}. Weights are recomputed before each Transformer layer while clean logits are cached once. No gradient update or change to the stored affine weight format is required.

\begin{algorithm}[t]
    \caption{\optimizer for a single linear layer}
    \label{alg:schurquant}
    \begin{algorithmic}[1]
        \Require Sufficient statistics $(G,C)$, reference weights $\Wref$,
        bit width $b$, group size $g$, refinement count $R$, and damping
        coefficient $\epsilon_{\mathrm{damp}}$
        \Ensure Quantized weights $W$

        \State $P \gets
        \bigl(G+\epsilon_{\mathrm{damp}}\,\operatorname{diag}(G)I\bigr)^{-1}$
        \State Initialize the working matrix $W \gets \Wref$

        \For{$t=0,1,\ldots,d_{\mathrm{in}}/g-1$}
            \State Partition
            $
            P =
            \begin{bmatrix}
                P_{cc} & P_{cr} \\
                P_{cr}^\top & P_{rr}
            \end{bmatrix}
            $
            according to the current chunk
            \State $S \gets P_{cc}^{-1}$
            \State $K \gets -P_{cr}^\top S$

            \State Compute $C_c^{\mathrm{eff}}$ and $C_r^{\mathrm{eff}}$
            \State $T \gets C_c^{\mathrm{eff}}-C_r^{\mathrm{eff}}K$

            \State Initialize $(Z,a,o)$ by row-wise min--max
            quantization of $\Wref_c$ to $b$ bits

            \For{$1,2,\ldots,R$}
                \State Fit $(a,o)$ by Eq.~\ref{eq:qparam-update}
                \State $Q \gets \operatorname{diag}(a)(Z-o\mathbf{1}^\top)$
                \State Sweep all entries of $Z$ using
                Eq.~\ref{eq:coordinate-objective}
            \EndFor

            \State Refit $(a,o)$; set $W_c\gets\operatorname{diag}(a)(Z-o\mathbf{1}^\top)$
            \State $P \gets P_{rr}-P_{cr}^\top S P_{cr}$
        \EndFor

        \State \Return $W$
    \end{algorithmic}
\end{algorithm}

\subsection{Complete algorithm}
\label{sec:algorithm}

Each chunk is initialized by row-wise min--max quantization of $\Wref_c$, then alternates the exact update of $(a,o)$ with coordinate descent over $Z$ for $R$ refinement steps, followed by a final parameter refit because the last sweep changes $Z$.
Each block update is non-increasing in $\mathcal{L}_{\mathrm{local}}$, and the finite code space implies convergence to a coordinate-wise fixed point. 
Unlike GPTQ, \optimizer needs no explicit suffix error-propagation step. 
The optimal response is already incorporated into $(S,T)$, and the next chunk's effective linear term reflects the fixed prefix. 
If $g=1$, $(a,o)$ are held fixed, and refinement is restricted to a single nearest-level decision, the procedure reduces to GPTQ's column-wise update. 
Algorithm~\ref{alg:schurquant} states the full procedure.

\section{Experiments}
\label{sec:experiments}

\subsection{Experimental setup}
\label{sec:experimental-setup}

\paragraph{Quantization.}
Unless stated otherwise, all weight-only experiments use group size $g=128$, a row-wise asymmetric quantizer, and float16 computation. We follow the standard GPTQ calibration protocol and quantize the linear attention and MLP projections inside Transformer blocks. Full quantization and calibration settings are provided in Appendix~\ref{app:quantization-details}.

\paragraph{Models.}
We evaluate Llama2-7B/13B~\citep{arxiv-2023-llama2}, Llama3.2-1B/3B and Llama3-8B~\citep{arxiv-2024-llama3}, and Qwen3-0.6B/4B/8B~\citep{arxiv-2025-qwen3}.
They span 0.6B--13B parameters and include distinct tokenizers and normalization designs.

\paragraph{Metrics.}
We evaluate perplexity (PPL) on the complete WikiText-2 test split~\citep{iclr-2017-wikitext} using windows of length 2,048:
\begin{equation}
    \mathrm{PPL}=\exp\!\left(-\frac1N\sum_{i=1}^{N}
    \log p(x_i\mid x_{<i})\right).
\end{equation}
We evaluate zero-shot accuracy on PIQA~\citep{aaai-2020-piqa}, ARC-Easy and ARC-Challenge~\citep{arxiv-2018-arc}, HellaSwag~\citep{acl-2019-hellaswag}, WinoGrande~\citep{aaai-2020-winogrande}, and BoolQ~\citep{naacl-2019-boolq}.
We use length-normalized accuracy for ARC-Easy, ARC-Challenge, and HellaSwag and standard accuracy otherwise, and report the unweighted mean across the six tasks.

\paragraph{Baselines and environment.}
GPTQ, QEP, FOEM, and \method are evaluated within the same quantization codebase and calibration protocol. 
Experiments use Python~3.12, PyTorch~2.11.0, Transformers~5.14.1, GPTQModel~7.3.4, Triton~3.6.0, and CUDA runtime~13.0.

\subsection{WikiText-2 perplexity}
\label{sec:ppl-results}

\begin{table}[t]
\caption{WikiText-2 test perplexity (lower is better). FP16 is the full-precision reference and SchQ denotes \method{}. For each model and bit-width, the best quantized result is bold.}
\label{tab:ppl}
\centering
\setlength{\tabcolsep}{3pt}
\resizebox{\textwidth}{!}{%
\begin{tabular}{l r rrrr rrrr rrrr}
\toprule
& & \multicolumn{4}{c}{2 bits} & \multicolumn{4}{c}{3 bits} & \multicolumn{4}{c}{4 bits}\\
\cmidrule(lr){3-6}\cmidrule(lr){7-10}\cmidrule(lr){11-14}
Model & FP16 & SchQ & GPTQ & FOEM & QEP & SchQ & GPTQ & FOEM & QEP & SchQ & GPTQ & FOEM & QEP\\
\midrule
Llama2-7B   & 5.46 & \textbf{10.54} & 37.63 & 68.73 & 29.84 & \textbf{6.07} & 6.32 & 6.38 & 6.39 & \textbf{5.59} & 5.62 & 5.64 & 5.71\\
Llama2-13B  & 4.88 & \textbf{8.28} & 16.69 & 15.42 & 13.50 & \textbf{5.33} & 5.42 & 5.43 & 5.43 & 4.99 & 4.98 & \textbf{4.97} & 4.98\\
Llama3.2-1B & 9.76 & \textbf{79.19} & 3432.88 & 6034.37 & 1248.04 & \textbf{13.25} & 15.41 & 68.22 & 15.26 & \textbf{10.42} & 10.63 & 43.25 & 10.70\\
Llama3.2-3B & 7.82 & \textbf{29.39} & 10549.22 & 4331.26 & 378.70 & \textbf{9.39} & 12.07 & 253.99 & 18.05 & \textbf{8.12} & 8.24 & 17.33 & 8.51\\
Llama3-8B   & 6.14 & \textbf{36.69} & 486.51 & 554.52 & 221.76 & \textbf{7.55} & 8.31 & 15.72 & 9.79 & \textbf{6.46} & 6.57 & 11.04 & 6.64\\
Qwen3-0.6B  & 20.96 & \textbf{74.72} & 12027.65 & 106866.61 & 659.24 & \textbf{26.90} & 50.53 & 39.16 & 35.39 & \textbf{22.63} & 26.22 & 24.13 & 23.77\\
Qwen3-4B    & 13.64 & \textbf{32.40} & 258.96 & 580.99 & 80.61 & \textbf{15.48} & 15.79 & 15.85 & 15.54 & 14.22 & \textbf{14.16} & 14.31 & 14.40\\
Qwen3-8B    & 9.72 & \textbf{29.48} & 77.13 & 165.42 & 57.47 & 11.16 & \textbf{10.92} & 11.66 & 11.37 & 10.02 & \textbf{9.91} & 10.03 & 10.17\\
\bottomrule
\end{tabular}}
\end{table}

Table~\ref{tab:ppl} summarizes WikiText-2 perplexity across all models and bit widths.
At 2 bits, \method{} improves substantially over GPTQ on all eight models. 
It prevents the extreme collapses observed for Llama3.2-1B (3432.88 to 79.19), Llama3.2-3B (10549.22 to 29.39), and Qwen3-0.6B (12027.65 to 74.72), and gives the lowest PPL among all methods on every model.
At 3 bits, \method{} is best on seven of the eight models; only Qwen3-8B favors GPTQ (10.92 versus 11.16).
At 4 bits, \method{} is best on five models, while FOEM is best on Llama2-13B and GPTQ is best on Qwen3-4B/8B.
As Section~\ref{sec:zero-shot-results} shows, this reversal is specific to perplexity.

\subsection{Zero-shot accuracy}
\label{sec:zero-shot-results}

\begin{table}[t]
\caption{Mean zero-shot accuracy (\%; higher is better) across the six evaluated tasks. For each model and bit-width, the best quantized result is bold.}
\label{tab:zeroshot}
\centering
\setlength{\tabcolsep}{3pt}
\resizebox{\textwidth}{!}{%
\begin{tabular}{l r rrrr rrrr rrrr}
\toprule
& &
\multicolumn{4}{c}{2 bits} & \multicolumn{4}{c}{3 bits} & \multicolumn{4}{c}{4 bits} \\
\cmidrule(lr){3-6}
\cmidrule(lr){7-10}
\cmidrule(lr){11-14}
Model & FP16
& SchQ & GPTQ & FOEM & QEP  & SchQ & GPTQ & FOEM & QEP & SchQ & GPTQ & FOEM & QEP \\
\midrule
Llama2-7B
& 70.45
& \textbf{56.58} & 41.38 & 39.47 & 43.73
& \textbf{68.13} & 67.03 & 67.04 & 67.53
& 69.54 & \textbf{70.07} & 69.89 & 69.93 \\

Llama2-13B
& 73.29
& \textbf{62.55} & 48.33 & 40.35 & 50.14
& 71.59 & \textbf{71.84} & 71.08 & 71.63
& \textbf{73.11} & 73.06 & 73.04 & 73.09 \\

Llama3.2-1B
& 60.35
& \textbf{45.05} & 37.60 & 36.45 & 37.07
& \textbf{55.49} & 53.94 & 52.07 & 53.51
& \textbf{59.28} & 59.19 & 57.88 & 58.56 \\

Llama3.2-3B
& 69.04
& \textbf{47.90} & 36.72 & 36.90 & 41.11
& \textbf{65.98} & 63.05 & 60.54 & 60.56
& 68.04 & \textbf{68.11} & 66.52 & 67.81 \\

Llama3-8B
& 74.61
& \textbf{48.53} & 38.04 & 37.20 & 37.73
& \textbf{72.20} & 66.15 & 66.72 & 68.22
& \textbf{74.27} & 72.94 & 73.03 & 73.27 \\

Qwen3-0.6B
& 54.19
& \textbf{44.19} & 36.63 & 36.33 & 37.50
& \textbf{49.82} & 44.43 & 46.04 & 45.75
& \textbf{52.43} & 48.37 & 48.25 & 49.76 \\

Qwen3-4B
& 71.17
& \textbf{51.50} & 36.77 & 36.36 & 40.31
& \textbf{65.79} & 65.53 & 62.82 & 65.44
& 70.19 & 69.50 & \textbf{70.39} & 69.00 \\

Qwen3-8B
& 74.09
& \textbf{50.31} & 38.53 & 36.95 & 41.81
& \textbf{71.89} & 69.56 & 68.92 & 70.94
& 73.43 & \textbf{73.60} & 72.70 & 73.47 \\

\midrule
Mean
& 68.40
& \textbf{50.83} & 39.25 & 37.50 & 41.18
& \textbf{65.11} & 62.69 & 61.90 & 62.95
& \textbf{67.54} & 66.86 & 66.46 & 66.86 \\
\bottomrule
\end{tabular}
}
\end{table}

Table~\ref{tab:zeroshot} reports the mean zero-shot accuracy across the six downstream tasks for all models and bit widths. 
The results reveal three distinct regimes across quantization precisions.
At 2 bits, \method{} is best on all eight models and averages 50.83\%, exceeding QEP by 9.65 pp and GPTQ by 11.58 pp.
Search error from greedy rounding is dominant on this coarse grid. 
At 3 bits, \method{} averages 65.11\%, 2.16 pp above QEP, and wins on seven of eight models.
On Llama3-8B, it exceeds GPTQ by 6.05 pp.
Notably, the 3-bit Qwen3-8B perplexity reversal in Table~\ref{tab:ppl} is flipped again in zero-shot accuracy. 
\method{} is 2.33 pp above GPTQ. 
At 4 bits, \method{} remains best on average at 67.54\%, but its 0.68-point advantage over the strongest baselines is much smaller than at lower precision, consistent with the diminishing difference between nearest rounding and a better discrete assignment on a finer grid.

\subsection{Optimizer component ablation}
\label{sec:optimizer-components}

Table~\ref{tab:optimizer-components} holds the 2-bit current-input self-reconstruction objective, code descent, initialization, $g=128$, $R=16$, windows, and seed fixed on Qwen3-4B. ``Raw'' holds the suffix fixed, ``Schur'' eliminates its continuous response. ``Fixed'' retains the initial grid, ``refit'' applies Eq.~\ref{eq:qparam-update}.

\begin{table}[t]
\caption{Controlled optimizer ablation on Qwen3-4B at 2 bits. $\Delta\mathcal L$ is the normalized layer-0 Q/K/V local-loss increase in percent.}
\label{tab:optimizer-components}
\centering
\small
\setlength{\tabcolsep}{5pt}
\begin{tabular}{llrr}
\toprule
Curvature & Grid & $\Delta\mathcal L$ (\%) $\downarrow$ & PPL $\downarrow$\\
\midrule
Raw & fixed & 1.645 & 2344.82\\
Raw & refit & 0.872 & 152.95\\
Schur & fixed & 1.165 & 324.81\\
Schur & refit & \textbf{0.550} & \textbf{86.64}\\
\bottomrule
\end{tabular}
\end{table}

With the grid fixed, Schur conditioning reduces the controlled loss from 1.645\% to 1.165\% and PPL from 2344.82 to 324.81. With the same refit on both curvatures, it reduces the loss from 0.872\% to 0.550\% (36.9\% relative) and PPL from 152.95 to 86.64. Thus, the gain is not attributable to coordinate descent or grid refitting alone.

\subsection{Objective ablation}
\label{sec:objective-ablation}

Table~\ref{tab:objective-ablation} varies the optimizer and sufficient statistics on Qwen3-4B. Rows 1--2 share the GPTQ objective. The remaining rows use quantized-prefix inputs and clean teacher outputs; residual-add rows extend the target to the sublayer output. Weighted rows use $\rho=8$ and Eq.~\ref{eq:effective-token-weight}, and $\lambda=1$ denotes the reference-weight anchor.

\begin{table}[t]
\caption{Objective ablation on Qwen3-4B. Mean is six-task zero-shot accuracy (\%).}
\label{tab:objective-ablation}
\centering
\setlength{\tabcolsep}{4pt}
\resizebox{\textwidth}{!}{%
\begin{tabular}{l rr rr rr}
\toprule
& \multicolumn{2}{c}{2 bits} & \multicolumn{2}{c}{3 bits} & \multicolumn{2}{c}{4 bits}\\
\cmidrule(lr){2-3}\cmidrule(lr){4-5}\cmidrule(lr){6-7}
Optimizer and objective & PPL $\downarrow$ & Mean $\uparrow$ & PPL $\downarrow$ & Mean $\uparrow$ & PPL $\downarrow$ & Mean $\uparrow$\\
\midrule
GPTQ optimizer + GPTQ objective & 258.96 & 36.77 & 15.79 & 65.53 & \textbf{14.16} & 69.50\\
\optimizer + GPTQ objective & 81.95 & 48.65 & 21.26 & 65.41 & 15.52 & 69.58\\
\optimizer + teacher, $\lambda=1$ & 75.01 & 42.97 & 16.23 & 65.13 & 14.29 & 69.06\\
\optimizer + teacher + weights, $\lambda=0$ & 64.81 & 43.62 & 18.92 & 63.77 & 15.69 & 69.02\\
\optimizer + teacher + weights, $\lambda=1$ & 67.70 & 46.20 & 16.07 & \textbf{66.83} & 14.31 & 70.00\\
\optimizer + teacher + residual-add target, $\lambda=1$ & 38.72 & 44.84 & 15.99 & 62.30 & 14.22 & 68.80\\
\optimizer + teacher + residual-add target + weights, $\lambda=0$ & 38.94 & 48.17 & 18.63 & 65.87 & 16.01 & 69.54\\
\method (full), $\lambda=1$ & \textbf{32.40} & \textbf{51.50} & \textbf{15.48} & 65.79 & 14.22 & \textbf{70.19}\\
\bottomrule
\end{tabular}}
\end{table}

With the GPTQ objective fixed, \optimizer changes 2-bit PPL from 258.96 to 81.95 and accuracy from 36.77\% to 48.65\% ($+11.88$ pp). At 3 and 4 bits, it worsens PPL while changing mean accuracy by $-0.12$ and $+0.08$ pp, respectively. For the projection-local teacher target, weighting improves mean accuracy at every precision, and adding the anchor on top of weighting gives a further $+2.58$, $+3.06$, and $+0.98$ pp at 2, 3, and 4 bits. With weights and $\lambda=1$ fixed, the residual-add target lowers PPL at all three precisions and changes mean accuracy by $+5.30$, $-1.04$, and $+0.19$ pp, respectively. The full configuration therefore gives the best 2-bit mean, the best 2/3-bit PPL, and the best 4-bit mean; the GPTQ optimizer gives the best 4-bit PPL, while the projection-local weighted objective remains best in 3-bit mean accuracy.

\subsection{Convergence with refinement steps}
\label{sec:refinement-ablation}

\begin{table}[t!]
\caption{WikiText-2 perplexity versus refinement count $R$ for 2-bit Llama2-7B under the fixed GPTQ objective. $R=0$ uses only min--max initialization.}
\label{tab:refinement}
\centering
\small
\begin{tabular}{lrrrrrrr}
\toprule
$R$ & 0 & 1 & 2 & 4 & 8 & 16 & 32\\
\midrule
PPL $\downarrow$ & 70.46 & 37.93 & 28.75 & \textbf{20.54} & 21.41 & 21.19 & 21.21\\
\bottomrule
\end{tabular}
\end{table}

Table~\ref{tab:refinement} reports WikiText-2 perplexity across different refinement counts $R$. 
Perplexity improves rapidly and saturates around $R=4$. 
Variation over $R\ge4$ is only 0.87, compared with a 49.92 improvement from $R=0$ to $R=4$. 
The local quadratic objective is monotone in $R$, whereas perplexity is not, again illustrating surrogate over-optimization. 
We use $R=16$ elsewhere because it was stable across models and bit widths.

\section{Conclusion}
\label{sec:conclusion}

\optimizer analytically eliminates the optimal continuous suffix response and optimizes each group under the resulting Schur-reduced quadratic, including a closed-form row-wise scale/zero-point refit. With the GPTQ objective fixed on 2-bit Qwen3-4B, replacing GPTQ's optimizer reduces PPL from 258.96 to 81.95 and raises mean accuracy by 11.88 pp.
The component ablation separately tests the conditional curvature. 
\method pairs this optimizer with quantized-prefix teacher reconstruction, reference-weight regularization, residual-add targets, and teacher-decision token weighting, reaching mean accuracy of 50.83\%, 65.11\%, and 67.54\% at 2, 3, and 4 bits across eight models.

\section{Limitations}
Our method has two main limitations. First, \method incurs higher offline quantization cost than existing PTQ methods due to repeated calibration and discrete refinement, limiting its scalability to larger models. Second, optimizing the layer-wise surrogate does not always translate to improved end-task performance, and our evaluation is limited to asymmetric quantization, leaving broader quantization settings for future work.

\subsubsection*{Reproducibility statement}
The weighted objective, token-weight construction, sufficient statistics, Schur recurrence, quantizer updates, and code-sweep complexity are specified in Sections~\ref{sec:problem-formulation}--\ref{sec:algorithm}. Section~\ref{sec:experiments} reports the models, calibration data, evaluation protocol, software environment, and all principal hyperparameters. Complete derivations are provided in Appendix~\ref{app:derivations}.

\subsubsection*{AI use statement}
Generative AI was utilized for minor tasks like grammar checking. 

\bibliography{ref}
\bibliographystyle{iclr2027_conference}

\appendix
\section{Calibration}
\label{sec:calibration}

The purpose of quantization is not to preserve weights in isolation, but to preserve the model's outputs. Let $E=\Wref-W$ be the weight error and let $X=[x_1,\ldots,x_N]$ contain calibration inputs. The total squared output error is
\begin{equation}
    \lVert EX\rVert_F^2
    =\sum_{k=1}^{N}\lVert Ex_k\rVert_2^2
    =\tr\!\left(E\,XX^\top E^\top\right).
    \label{eq:calibration-curvature}
\end{equation}
The Gram matrix $XX^\top$ weights errors by the activation distribution: errors in strongly activated directions matter more than equally large errors in nearly inactive directions. Data-free round-to-nearest minimizes $\lVert E\rVert_F^2$, which implicitly replaces $XX^\top$ by the identity and treats every direction equally. Calibration estimates the missing curvature from a small sample of activations. Standard calibration assigns equal explicit mass to all token columns. With nonnegative token weights $\omega_i$, the same derivation becomes
\begin{equation}
    \sum_{i=1}^{N}\omega_i\lVert Ex_i\rVert_2^2
    =\tr\!\left(E X\Omega X^\top E^\top\right),
    \qquad \Omega=\operatorname{diag}(\omega_1,\ldots,\omega_N).
    \label{eq:weighted-calibration-curvature}
\end{equation}
Thus token weighting changes the empirical activation distribution seen by the quantizer without changing the quadratic form of the problem. This weighted curvature is the sufficient-statistics modification used by teacher-decision weighting in Section~\ref{sec:align-objective}. 
The residual-add target independently modifies the weighted cross-Gram $Y\Omega X^\top$. 
The unweighted special cases recover the Hessians used by GPTQ~\citep{iclr-2023-gptq} and QEP~\citep{neurips-2025-qep}.

\section{Detailed Derivations}
\label{app:derivations}

\subsection{Proof of Proposition~\ref{prop:schur}}
\label{app:proof-schur}

Terms containing only the fixed prefix $W_F$ are constant. 
Expanding the quadratic form in Eq.~\ref{eq:quadratic-objective}, the cross terms involving $W_F$ and $W_c$ are
\begin{equation}
    \frac12\tr\!\left(W_cG_{cF}W_F^\top+W_FG_{Fc}W_c^\top\right)
    =\tr(W_FG_{Fc}W_c^\top),
\end{equation}
where symmetry gives $G_{cF}=G_{Fc}^\top$. Combining this expression with $-\tr(C_cW_c^\top)$ yields the effective linear coefficient $C_c^{\mathrm{eff}}=C_c-W_FG_{Fc}$. The same argument gives $C_r^{\mathrm{eff}}=C_r-W_FG_{Fr}$ for the suffix.

Collecting all terms that depend on $(W_c,W_r)$ gives
\begin{align}
    \mathcal{L}(W_c,W_r)
    ={}&\frac12\tr\!\left(
    \begin{bmatrix}W_c&W_r\end{bmatrix}
    \begin{bmatrix}G_{cc}&G_{cr}\\G_{rc}&G_{rr}\end{bmatrix}
    \begin{bmatrix}W_c&W_r\end{bmatrix}^{\!\top}\right)\nonumber\\
    &-\tr(C_c^{\mathrm{eff}}W_c^\top)
     -\tr(C_r^{\mathrm{eff}}W_r^\top)+\mathrm{const}.
    \label{eq:appendix-block-objective}
\end{align}
Holding $W_c$ fixed and differentiating with respect to $W_r$ gives
\begin{equation}
    W_cG_{cr}+W_rG_{rr}-C_r^{\mathrm{eff}}=0,
\end{equation}
so the optimal continuous suffix response is
\begin{equation}
    W_r^\star(W_c)=\left(C_r^{\mathrm{eff}}-W_cG_{cr}\right)G_{rr}^{-1}.
    \label{eq:appendix-suffix-response}
\end{equation}
Let $U=C_r^{\mathrm{eff}}-W_cG_{cr}$. The suffix-dependent terms can be completed to a square
\begin{align}
    &\frac12\tr(W_rG_{rr}W_r^\top)-\tr(UW_r^\top)\nonumber\\
    &\quad=\frac12\tr\!\left((W_r-UG_{rr}^{-1})G_{rr}
       (W_r-UG_{rr}^{-1})^\top\right)
       -\frac12\tr(UG_{rr}^{-1}U^\top).
\end{align}
The first term vanishes at Eq.~\ref{eq:appendix-suffix-response}. Expanding the second gives
\begin{align}
    -\frac12\tr(UG_{rr}^{-1}U^\top)
    ={}&-\frac12\tr(C_r^{\mathrm{eff}}G_{rr}^{-1}C_r^{\mathrm{eff}\top})\nonumber\\
      &+\tr(C_r^{\mathrm{eff}}G_{rr}^{-1}G_{rc}W_c^\top)\nonumber\\
      &-\frac12\tr(W_cG_{cr}G_{rr}^{-1}G_{rc}W_c^\top).
\end{align}
The first term is constant in $W_c$. Combining the remaining terms with Eq.~\ref{eq:appendix-block-objective} produces the curvature $S$ and the linear coefficient $T$ of Eq.~\ref{eq:local-curvature}, proving Proposition~\ref{prop:schur}. \hfill$\square$

\subsection{Proof of Proposition~\ref{prop:qparam}}
\label{app:proof-qparam}

For a fixed zero-point $o$, substitute $q_i=a_i u_i(o)$ into Eq.~\ref{eq:rowwise-objective}:
\begin{equation}
    \mathcal{L}_i(a_i;o)
    =\frac12A_i(o)a_i^2-d_i(o)a_i.
\end{equation}
When $A_i(o)>0$, differentiating gives the unconstrained optimum
$d_i(o)/A_i(o)$; projecting it onto $a_i\geq\epsilon_a$ gives
Eq.~\ref{eq:qparam-update}. Enumerating the finite set of $b$-bit zero-points
and selecting the smallest substituted objective is globally exact for fixed
codes. \hfill$\square$

\section{Quantization runtime}
\label{sec:quantization-runtime}

Table~\ref{tab:runtime} reports measured end-to-end wall-clock, including statistics collection and discrete optimization. \method performs $R=16$ sweeps and one current full-model calibration pass before each Transformer layer in addition to the clean-cache pass. It is therefore an offline accuracy--quantization-time trade-off, not an inference-speed result.

\begin{table}[t]
\caption{End-to-end quantization time in minutes on one NVIDIA H100 PCIe GPU.}
\label{tab:runtime}
\centering
\small
\setlength{\tabcolsep}{4pt}
\begin{tabular}{llrrr}
\toprule
Model & Method & 2 bit & 3 bit & 4 bit\\
\midrule
\multirow{4}{*}{Qwen3-4B}
& GPTQ & 5.5 & 6.9 & 5.5\\
& FOEM & 6.4 & 7.6 & 6.6\\
& QEP & 55.6 & 71.2 & 69.4\\
& \method & 253.7 & 254.0 & 282.5\\
\midrule
\multirow{4}{*}{Llama2-7B}
& GPTQ & 7.1 & 8.4 & 7.1\\
& FOEM & 8.2 & 9.1 & 8.0\\
& QEP & 38.6 & 54.1 & 59.3\\
& \method & 224.2 & 220.9 & 235.6\\
\bottomrule
\end{tabular}
\end{table}

At 2 bits, quantization with \method takes 45.8$\times$/31.8$\times$ as long as GPTQ and 4.6$\times$/5.8$\times$ QEP on Qwen3-4B/Llama2-7B; the 3/4-bit GPTQ ratios are 37.0/51.1$\times$ and 26.5/33.0$\times$. Instrumentation now separates statistics and optimizer time and peak memory. Dense evaluator checkpoints are not packed-size evidence.

\section{Quantization setup} 
\label{app:quantization-details} 

Unless stated otherwise, all weight-only experiments use group size $g=128$, a row-wise asymmetric quantizer, $R=16$ refinement steps, and float16 computation. 
We draw 128 random windows of length 2,048 from long documents in the C4 training split~\citep{jmlr-2020-c4}, following the standard calibration protocol of GPTQ. 
The final \method{} configuration constructs the top-mismatch weights from all 128 calibration windows before quantizing each Transformer layer. 
We use mismatch multiplier $\rho=8$ and the detached projection-local residual normalization in Eq.~\ref{eq:effective-token-weight} with residual floor $\epsilon=10^{-8}$. 
We quantize the linear attention and MLP projections inside Transformer blocks while retaining the embeddings, language-model head, and normalization parameters in their original precision.

\end{document}